\documentclass[runningheads]{llncs}
\usepackage{graphicx}
\usepackage[round]{natbib}
\usepackage[utf8x]{inputenc}
\usepackage[T1]{fontenc}
\usepackage{setspace}
\usepackage{hyperref} 
\usepackage{amsmath,amssymb}
\usepackage{algorithmic} 
\usepackage[ruled,vlined]{algorithm2e} 

\usepackage{changepage} 
\usepackage{bm} 
\usepackage{nomencl} 
\usepackage{xfp} 
\usepackage[aboveskip=1pt,labelfont=bf,labelsep=period,singlelinecheck=off]{caption}
\makenomenclature
\usepackage{geometry}
\usepackage{subcaption} 
\usepackage{microtype} 
\usepackage{fancyhdr} 
\usepackage{ragged2e}

\begin{document}
\title{Reward prediction for representation learning \\ and reward shaping
}
%
%
\author{Hlynur Davíð Hlynsson\and
Laurenz Wiskott}
\authorrunning{Hlynsson and Wiskott}
%
\institute{Ruhr-Universität Bochum, 44801 Bochum, Germany}
\maketitle              
\begin{abstract}
One of the fundamental challenges in reinforcement learning (RL) is the one of data efficiency: modern algorithms require a very large number of training samples, especially compared to humans, for solving environments with high-dimensional observations. The severity of this problem is increased when the reward signal is sparse. In this work, we propose learning a state representation in a self-supervised manner for reward prediction. The reward predictor learns to estimate either a raw or a smoothed version of the true reward signal in environment with a single, terminating, goal state. We augment the training of out-of-the-box RL agents by shaping the reward using our reward predictor during policy learning. Using our representation for preprocessing high-dimensional observations, as well as using the predictor for reward shaping, is shown to significantly enhance Actor Critic using Kronecker-factored Trust Region and Proximal Policy Optimization in single-goal environments with visual inputs.

\end{abstract}
\section{Introduction}

Deep learning has enjoyed great success in recent years with a fortunate combination of increasing computational capabilities and advances in algorithm design in addition to an ample offering of flexible software ecosystems. In particular, reinforcement learning (RL), which is the discipline of machine learning  concerned with general goal-directed behavior, has shown great promise since DeepMind combined the principles of RL with deep learning to achieve human-like skill on Atari video games \citep{mnih2013playing}. Since then, the list of games where machine triumphs over man grows longer with the addition of algorithms surpassing human capabilities in games such as Go, \citep{silver2016mastering}, Poker \citep{brown2019superhuman} and a restricted version of DOTA 2 \citep{berner2019dota}.

Even though the dominance of humans is being tested by RL agents on numerous fronts, there are still great difficulties for the field to overcome. For instance, the data that is required for algorithms to reach human performance is on a far larger scale than that needed of humans. Furthermore, the general intelligence of humans remains unchallenged. Even though an RL agent has reached superhuman performance in one field, its performance is usually poor when it is tested in new areas.

The study of methods to overcome the problem of the data-efficiency and transferability of RL agents, in environments where the agent must reach a single goal, is the focal point of this work. We consider a simple way of learning a state representation by predicting either a raw or a smoothed version of a sparse reward yielded by an environment. The two objectives, of learning a representation and predicting the reward, are directly connected as we train a deep neural network for the prediction, and the hidden layers of this network learn our reward-predictive representation of the input data. 

The reward signal is created by collecting data from a relatively low number of initial episodes, using controller that acts randomly. The representation is then extracted from an intermediate layer of the prediction model and re-used as general preprocessing for RL agents, to  reduce the dimensionality of visual inputs. The agent processes inputs corresponding to its current state as well as the desired end state, which is analogous to mentally visualizing a goal before attempting to reach it. This general approach of relying on state representations, that are learned to predict the reward rather than maximizing it, has been motivated in the literature \citep{lehnert2020reward} and we show that our representation is well-suited for single-goal environments. 

We also investigate the effectiveness of augmenting the reward for RL agents, when the reward is sparse,  with a novel problem-agnostic reward shaping technique. The reward predictor, that is used to train our representation, is not only used as a part of an auxiliary loss function to learn a representation, but it is also used during training to encourage the agent to move closer to a goal location. Similar to advantage functions in the reinforcement learning literature \citep{schulman2015high}, given the trained reward predictor, the agent receives an additional reward signal if it moves from states with a low predicted reward to states with a higher predicted reward. We find this reward augmentation to be beneficial for our test environment that has the largest state-space.

\section{Background}

\subsection{Markov decision processes}
A partially-observable Markov decision process (POMDP) is a tuple

\begin{equation} \label{eq:pomdp}
(\mathcal{S}, \mathcal{A}, \mathcal{P}, \mathcal{R}, \mathbb{P}(s_0), \Omega, O, \gamma)
\end{equation}

\noindent which we will also refer to as the \textit{environment}. The tuple is made up of the following elements:
\begin{itemize}
    \item[$\mathcal{S}$:] The state space defines the possible configurations of the environment 
    \item[$\mathcal{A}$:] The action space describes how the agent is able to interact with the environment
    \item[$\mathcal{P}$:] The transition function $\mathcal{P}: \mathcal{S} \times \mathcal{A} \rightarrow \mathbb{P}(\mathcal{S})$ which dictates the effects of different actions in different states
    \item[$\mathcal{R}$:] The reward function $\mathcal{R}:  \mathcal{S} \times \mathcal{A} \times \mathcal{S} \rightarrow \mathbb{R}$ determines the immediate reward given to the agent for transitioning between any two states with any action
    \item[$\mathbb{P}(s_0)$:] The initial state distribution
    \item[$\Omega$:] The observation space defines the aspects of the environment that the agent can perceive 
    \item[$O$:] The observation function $O: \mathcal{S} \times \mathcal{A} \rightarrow \mathbb{P}(\Omega)$ decides what subset of the environment the agent receives after acting in a given state
    
    \item[$\gamma$:] The reward discount factor
\end{itemize}
 
The environment starts in a state drawn from $\mathbb{P}(s_0)$, from which the agent interacts sequentially with the environment by choosing an action $a_t$, chosen from the action space $\mathcal{A}$, at time steps $t$. The agent receives an observation $O_t$ and a reward $r_t$ after each action. 

A discount factor $\gamma \in (0, 1)$ is usually included in the definition of POMDPs and it comes into play in the optimization function of the agent. Namely, the objective of an RL agent is to learn a \textit{policy} $\pi$, which determines the behavior of the agent in the environment by mapping states to a probability distribution over $\mathcal{A}$, $\pi(a, s) = \mathbb{P}(a_t = a | s_t = s)$. The policy should maximize the expected discounted future sum of rewards, or the expected \textit{return}, where the return is defined as

\begin{equation} \label{eq:ret2}
R_\Sigma = \sum_{t=0}^\infty \gamma^t r_t
\end{equation}

The expectation of the return (Eq. \ref{eq:ret2}), given a policy $\pi$ and an initial state $s_0 = s$, is defined as the \textit{value function}

\begin{equation} \label{eq:vf2}
 V_\pi(s) = \mathbb{E}\left[ R_\Sigma | s_0 = s, \pi\right] =  \mathbb{E}\left[ \sum_{t=0}^\infty \gamma^t r_t | s_0 = s, \pi\right] 
 \end{equation}

There is at least one \textit{optimal policy} $\pi^*$ that is better than or equal to others: $V_{\pi^*}(s) \geq V_{\pi'}(s)$ for all states $s$ and all other policies $\pi'$. Model-free RL methods learn the optimal policy or a value function directly from experience without attempting to approximate the dynamics of the environment. Model-based RL algorithms learn the optimal policy $\pi^*$ by first estimating the transition function $\Tilde{\mathcal{P}}\approx \mathcal{P}$ and the reward function $\Tilde{\mathcal{R}}\approx \mathcal{R}$, the environment \textit{dynamics} or \textit{world model}. These functions are learned in a supervised fashion from a data set of observed transitions, $\mathcal{D} = (s_t, a_t, r_t, s_{t+1})$. The estimators making up a world model can be used in multiple different ways, depending on the algorithm, to derive the optimal policy.

\subsection{Reward shaping}

Sparse rewards in environments is a common problem for reinforcement learning agents. The agent's goal is to associate its inputs with actions that lead to high rewards, which can be a lengthy process if the agent only rarely experiences positive or negative rewards. 

Reward shaping \citep{mataric1994reward, ng1999policy, brys2015policy} is a popular method of modifying the reward function of an MDP to speed up learning. It is useful for environments with sparse rewards to augment the training of the agent but skillful applications of reward shaping can in principle aid the optimization for any environment -- although the efficacy of the reward shaping is highly dependent on the details of the implementation \citep{clark2016faulty}. Reward shaping has been shown to be useful in the last few years for complex video game environments, such as real-time strategy games \citep{efthymiadis2013using} and platformers \citep{brys2014multi} and it also been combined with deep neural networks to improve agents in first person shooting games \citep{lample2017playing}.

To illustrate, consider learning a policy for car racing. If the goal is train an agent to drive optimally, then supplying it with a positive reward for reaching the finish line is in theory sufficient, as it is equipped to learn the necessary representation to achieve this in the long run. However, if it is punished for actions that are never beneficial, for instance crashing into walls, it will prioritize learning to avoid such situtations, allowing it to explore more promising parts of the state space. 

Furthermore, just reaching the goal is insufficient when the competition is fierce. To make sure that we have a winning racer on our hands, a small negative reward can be introduced to urge the agent to reach the finish line quickly. Note that the details of the reward shaping in this example requires expert intervention from a designer who is familiar with the environment. It would be more generally useful if the reward shaping would be autonomously learned, just as the policy of the agent, as we propose to do in this work.

\subsection{Reward-predictive vs. reward-maximizing representations}
\label{predvsmax}
\begin{figure}[ht!]
  \begin{center}
        \includegraphics[width=0.95\linewidth]{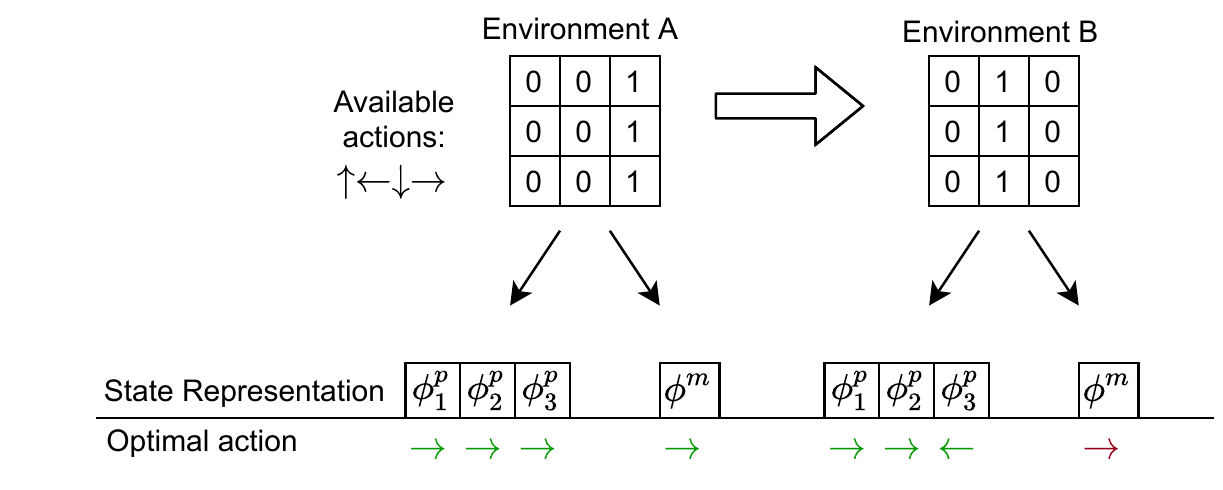}
  \caption[Illustration: Reward-maximizing vs. reward-predictive representations]{\textbf{Reward-maximizing vs. reward-predictive representations}. In this grid world example, the agent starts the episode at a random location and can move up, down, left and right. The episode ends with a reward of 1 and terminates when the agent reaches the rightmost column. Both the reward-predictive representation and reward-maximizing representation $\phi^p$ and $\phi^m$, respectively, are useful for learning the optimal policy in Environment A. The reward-predictive representation $\phi^p$ collapses each column into a single state to predict the discounted future reward. The reward-maximizing representation $\phi^m$ makes no such distinction as moving right is the optimal action in any state. It is a different story if the representations are transfered to Environment B, where reaching the middle column is now the goal. The representation $\phi^p$ can be reused and the optimal policy is found if agent now takes a step left in $\phi_3^p$. However, the representation $\phi^m$  is unable to discriminate between the different states and is useless for determining the optimal policy.}
  \label{fig:transferrep}
  \end{center}
\end{figure}

 \cite{lehnert2020reward} make the distinction between \textit{reward-maximizing} representations are \textit{reward-predictive} representations. They argue how reward-maximizing representations can transfer poorly to new environments while reward-predictive representations generalize successfully. Take the simple grid world navigation environments in Fig. \ref{fig:transferrep}, for example. The agent starts at a random tile in the grid and gets a reward of +1 by reaching the rightmost column in Environment A or by reaching the middle column in Environment B. 
 The state space in Environment A can be compressed from the $3 \times 3$ grid to a vector of length 3, $[\phi_1^p, \phi_2^p, \phi_3^p]$ of reward-predictive representations. To predict the discounted reward, it suffices to describe the agent's state with $\phi_j^p$ if it is in the $j$th row. 
 
 The reward-maximizing representation for Environment A is much simpler: the whole state space can be collapsed to a single element $\phi^m$, with the optimal policy of always moving to the right. 
 If these representations are kept, then the reward-predictive representation $\phi^p$ is informative enough for a RL agent to learn how to solve Enviroment B. The reward-maximizing representation $\phi^m$ has discarded too many details of the environment to be useful for solving this new environment.

\subsection{Successor features}
The \textit{successor representation} algorithm learns two functions: the expected reward $R_\pi^{\text{SF}}$ received after transitioning into a state $s$, as well as the matrix $M_\pi^{\text{SF}}$ of discounted expected future occupancy of each state,  assuming that the agent starts in a given state and follows a particular policy $\pi$.  Knowing the quantities $R_\pi^{\text{SF}}$ and $M_\pi^{\text{SF}}$ allows us to rewrite the value function:

 \begin{equation} \label{eq:sr0}
V_{\pi}(s) = \mathbb{E}_{s'}\left[ R_\pi^{\text{SF}}(s) M_\pi^{\text{SF}}(s, s')\right]\end{equation}

The motivation for this algorithm is that it combines the speed of model-free methods, by enabling fast computations of the value function, with the flexibility of model-based methods for environments with changing reward contingencies. 

This method is made for small, discrete spaces but it has been generalized for continuous spaces with so-called feature-based successor representations, or \textit{successor features} (SFs) \citep{barreto2016successor}.  The SF algorithm similarly calculates the discounted expected representation of future states, given the agent takes the action $a$ in the state $s$ and follows a policy $\pi$:

\begin{equation} \label{eq:sf}
\psi_{\pi}(s, a) = \mathbb{E}_\pi \left[\sum_{t=0}^\infty \gamma^{t-1} \phi_{t+1} | s_t = s, a_t = a\right]\end{equation}





\noindent where $\phi$ is some state representation. Both the SF $\psi$ and the representation $\phi$ can be deep neural networks.

\section{Related work}

 \subsection{Reward-predictive representations } 
 \cite{lehnert2020reward} compare successor features (SFs) to a nonparametric Bayesian predictor that is trained to learn transition and reward tables for the environment, either with a reward-maximizing or a reward-predictive loss function. \cite{lehnert2020successor} prove under what conditions successor features (SFs) are either \textit{reward-predictive} or \textit{reward-maximizing} (see distinction in Section \ref{predvsmax}). They also show that SFs work succesfully for transfer learning between environments with changing reward functions and unchanged transition functions, but they generalize poorly between environments where the transition function changes.
 
Our work is distinct from the reward-predictive methods that they compare as our representation does need to calculate expected future state occupancy, as is the case for SFs. Our method scales better for more complicated state-spaces because we do not tabulate the states, as they do with their Bayesian model, but learn arbitrary continuous features of high-dimensional input data. In addition to that, learning our reward predictor is not only a "surrogate" objective function as we use it for reward shaping as well.

\subsection{Reward shaping}

The advantages of reward shaping are well understood in the literature \citep{mataric1994reward}. A recent trend in RL research is the study of methods that can learn the reward shaping function in an automatic manner, without the need of (often faulty) human intervention. 

\cite{marashi2012automatic} assume that the environment can be expressed as a graph and that this graph formulation is known. Under these strong assumptions, they perform graph analysis to extract a reward shaping function. More recently, \cite{zou2019reward} propose a meta-learning algorithm for potential-based automatic reward shaping. Our approach is different from previous work as we assume no knowledge about the environment and train a simple predictor to approximate (potentially smoothed) rewards, which is then used to construct a potential-based reward shaping function.






\subsection{Goal-conditioned reinforcement learning } \cite{kaelbling1993learning} studied  environments with multiple goals and small state-spaces. Their problem setting has an agent with the objective of reaching a known but dynamically changing goal in the fewest number of moves. The observation space is of a low enough dimension for dynamic programming to be satisfactory in their case. \cite{schaul2015universal} introduce the  Universal Value Function Approximators and tackle environments of larger dimensions by  learning a value function neural network approximator that accepts both the current state and a goal state as the inputs. In a similar vein, \cite{pathak2018zero} learn a policy that is given a current state and a goal state and outputs an action that bridges the gap between them. These methods simultaneously learn the representation of the visual input as well learning the policy. In contrast, we learn a reward-predictive representation separately from maximizing the reward, in a self-supervised manner.

\section{Approach}
\label{chaptern_method}

In this section, we explain our approach mathematically. Intuitively, we train a deep neural network to predict either a raw or a smoothed reward signal from a single-goal environment. The output of an intermediate layer in the network is then extracted as the representation\footnote{For example, by simply removing the top layers of the network.}. The usage of this representation, on its own, is motivated by the illustration in Fig. \ref{predvsmax} and the surrounding text. The full reward predictor network is then used for reward shaping by rewarding the agent for moving from lower predicted values toward higher predicted values of the network.

\subsection{Learning the representation}

Suppose that $f_\theta: \mathbb{R}^c \rightarrow [0, 1]$ is a differentiable function parameterized by $\theta$ and $c$ is a positive integer. We use $f_\theta$ to approximate the discounted reward in a POMDP with a sparse reward: the agent receives a reward of 0 for each time step except when it reaches a goal location, at which point it receives a positive reward and the episode terminates.

Given an experience buffer $\mathcal{D} = $ $(s_t, a_t, r_t, s_{t+1})$, we create a new data set $\mathcal{D}^* = \left(s_t, a_t, r^*_t, s_{t+1} \right)$. The new rewards are calculated according to the equation  

\begin{equation} \label{eq:rstar}
r^*_t = \gamma^m r_{t+m}\end{equation}

\noindent where $\gamma \in [0, 1]$ is a discount factor and $M>m>0$ is the difference between $t$ and the time step index of the final transition in that episode, for some maximum time horizon $M$. Throughout our experiments, we keep the value of the discount factor equal to $0.99$ and we compare setting the maximum time horizon equal to $64$ or equal to $1$ (raw reward prediction).

\begin{figure}[ht]
\begin{center}
  \includegraphics[width=.85\linewidth]{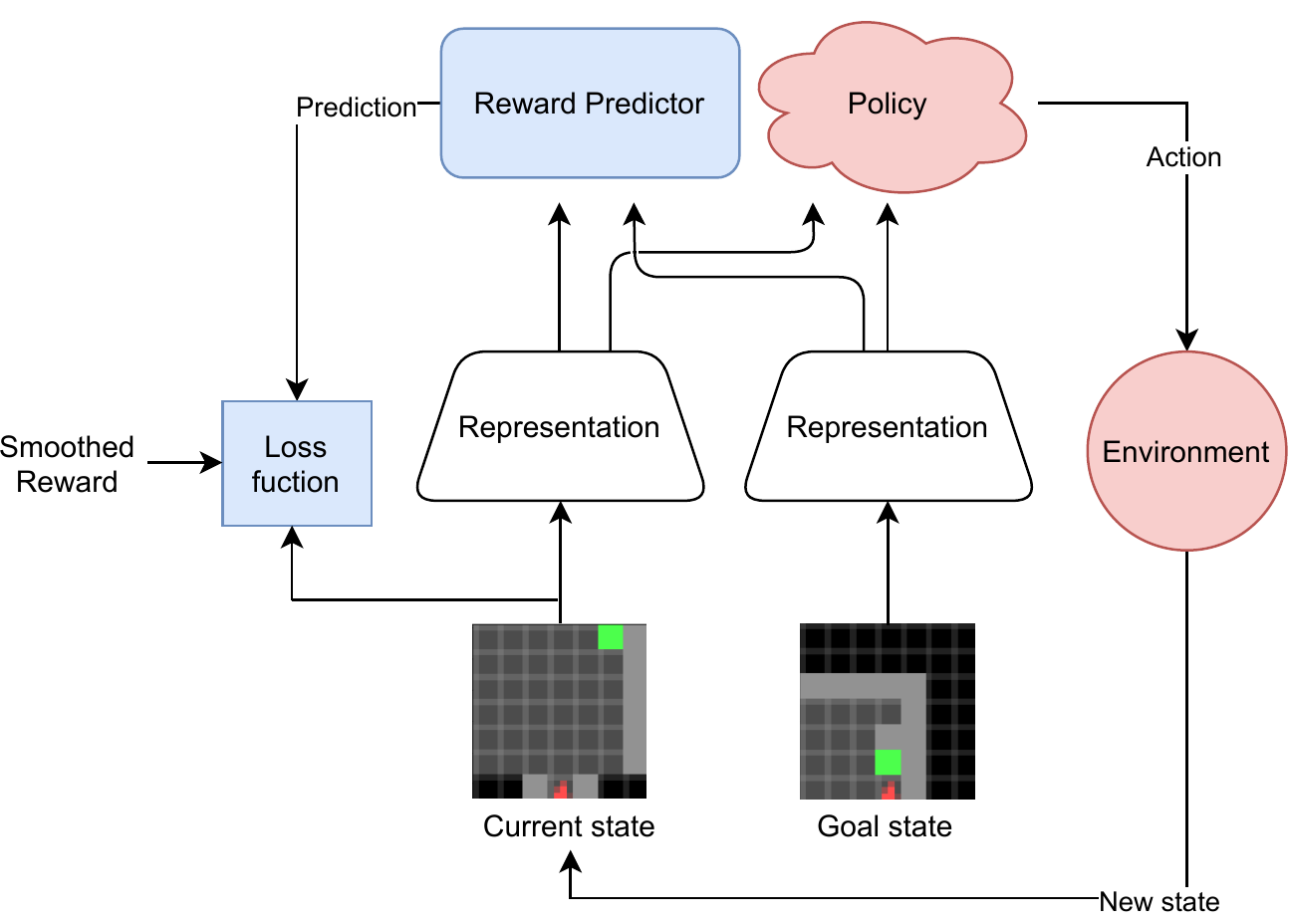}
  \caption[Illustration: Learning and using the representation]{\textbf{Learning and using the representation.} Our representation and reward predictor is trained with the elements highlighted in blue. The trained representation is then used for dimensionality reduction for an RL agent, that interacts with the environment as indicated by the elements highlighted in red. }
  \label{fig:boat5}
\end{center}
\end{figure}

Assume that our differentiable representation function $\phi: \mathbb{R}^d \rightarrow \mathbb{R}^c$ is parameterized by $\theta'$ and maps the $d$-dimensional raw observation of the POMDP to the $c$-dimensional feature vector. We train the representation for the discounted-reward prediction by minimizing the loss function

\begin{equation}
\mathcal{L}(f_\theta [ \phi_{\theta'} (s_{t+1})], r^*_t) =  \left( r^*_t - f_\theta [ \phi_{\theta'} (s_{t+1})] \right)^2
    \label{discounted_loss}
\end{equation}

\noindent with respect to the parameters $\theta$ of $f$ and the parameters $\theta'$ of $\phi$ over the whole data set $\mathcal{D}^*$. See Fig. \ref{fig:boat5} for a conceptual overview of our representation learning. 

\subsection{Reward shaping}

\cite{ng1999policy} define a reward shaping function $F$ as  \textit{potential-based} if there exists a function $f: \mathcal{S} \rightarrow \mathbb{R}$ such that for all states $s, s' \in \mathcal{S}$ the following equation holds:

\begin{equation}
F(s, a, s') = \gamma f(s') - f(s)  
    \label{potential_based}
\end{equation}

\noindent and $\gamma$ is the MDP's discount factor. They prove for single-goal environments that every optimal policy for the MDP $M = (\mathcal{S}, \mathcal{A}, \mathcal{P}, \mathcal{R}, \mathbb{P}(s_0), \gamma)$ is also optimal for its reward-shaped counterpart $M' = (\mathcal{S}, \mathcal{A}, \mathcal{P}, \mathcal{R}+F, \mathbb{P}(s_0), \gamma)$, and vice versa. They also show, for a given state space $\mathcal{S}$ and action space $\mathcal{A}$, that if $F$ is not a potential-based, then there exist a transition $T$ and a reward function $R$ such that no optimal policy in $M'$ is optimal in $M$.

Assume that we have an environment where an agent is tasked with reaching a goal state $g$. That is, for a given distance function $d: \mathcal{S} \times \mathcal{S} \rightarrow \mathbb{R}^+$ the agent receives a reward of 1 if it is close enough to the goal location, $d(s, g) \leq \delta$, for some reward threshold $\delta \in \mathbb{R}^+$. Otherwise, it receives a reward of 0.

The distance $d$ between the agent's location $s$ and the goal location $g$ can be a useful value to calculate in the design of a reward shaping function

\begin{equation}
F(s, a, s') = \begin{cases} 1 & \text{if } d(s', g) \leq \delta \\
                      -d(s', g)                                    & \text{otherwise}      %
        \end{cases}    
\label{negdist}
\end{equation}

This is sensitive to the form of $d$, as the agent could get stuck in local optima before coming close to the goal, i.e. if it would have to move through a region with a large $d(s, g)$ before it can globally minimize it. This could for example be the case in a maze environment if $d$ is the Euclidean distance between the $(x, y)$ coordinates of the agent's location and the goal location and there is a wall between the agent and the goal.  \cite{trott2019keeping} propose to solve this by incorporating the potential local optima in the reward shaping as function as so-called "anti-goals" $\Bar{g}$ to be avoided

\begin{equation}
F(s, a, s') = \begin{cases} 1 & \text{if } d(s', g) \leq \delta \\
                      \min [0, -d(s', g) + d(s, \Bar{g})]                               & \text{otherwise}      %
        \end{cases}    
\label{antigoals}
\end{equation}

These states can be hand-picked by domain experts. However, adding anti-goals like this could iteratively introduce even more local optima and a solution to the original problem is not guaranteed.

It is generally not true that the distance function $d$ and all the variables needed to calculate it, such as the coordinates of the agent and the goal in a maze, are available to the agent. Even if $d$ were computable, using it naïvely can bring about its own problems as was alluded to above. 

We argue that instead of using $d$ in Equation~\ref{negdist}, it would be better to measure the distance between the agent and the goal in terms of how many actions the agent has to take such that the goal is reached. This function is not assumed to be given, but it can be estimated as the agent is being trained on the environment, for instance by optimizing Equation~\ref{discounted_loss}.

Additionally, we would like our reward shaping function to be potential-based (Equation \ref{potential_based}) to reap the theoretical advantages. Thus, we propose a potential-based reward shaping function based on the discounted reward predictor


\begin{equation}
\begin{split}
F(s, a, s') & = \left( \gamma f_\theta ( \phi_{\theta'} [s']) - f_\theta ( \phi_{\theta'} [s]) \right)  (H - I) / H \\
 & = \gamma \left( f_\theta ( \phi_{\theta'} [s']) (H - I) / H \right) - f_\theta ( \phi_{\theta'} [s]) (H - I) / H \\
 & = \gamma f^*(s') - f^*(s)
\end{split}
\label{ourshaped}
\end{equation}

\noindent where $f^* =  f_\theta ( \phi_{\theta'} [s']) (H - I) / H $, $f_\theta$ is the reward predictor and $\phi_{\theta'}$ is our representation from the previous section. Note that both $f_\theta$ and $\phi_{\theta'}$ are assumed to be fully trained before the policy of the agent is trained, for example using data gathered by a random policy, but they can in principle also be updated as the policy is being learned. The factor $(H - I) / H$ scales down the intensity of the reward shaping while  $I \in \mathbb{N}^+$ is the number of episodes that the agent has experienced and $H \in \mathbb{N}^+$ is the maximum number of episodes where the agent is trained using reward shaping.  The strength of the reward shaping is the highest in the beginning to counteract potentially adverse effects of errors in the reward predictor. It is also more important to incentivize moving toward the general direction of the goal in the early stages of learning, after which the un-augmented reward signal of the environment is allowed to "speak for itself" and guide the learning of the agent precisely toward the goal.




\section{Methodology and implementation} \label{sec:exp}

\subsection{Environment} \label{subsec:env}

\begin{figure}
\centering
\begin{subfigure}[b]{0.75\textwidth}
   \includegraphics[width=1\linewidth]{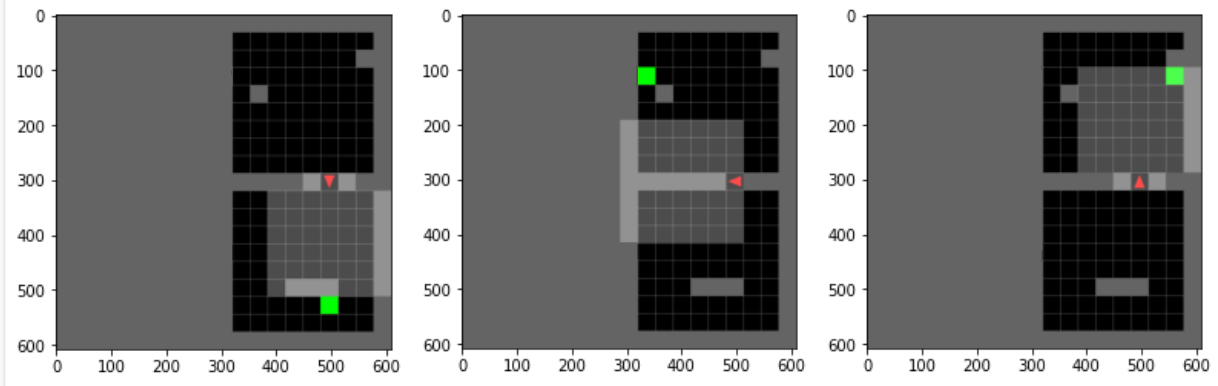}
   \caption{Full world states. }
   \label{fig:Nw1b} 
\end{subfigure}
\begin{subfigure}[b]{0.75 \textwidth}
   \includegraphics[width=1\linewidth]{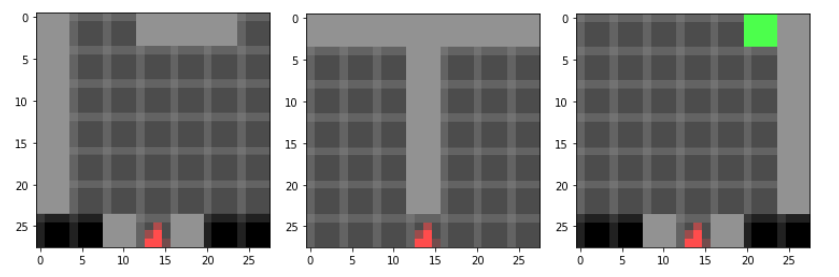}
   \caption{Agent's point of view.}
   \label{fig:Nw2obses}
\end{subfigure}
\begin{subfigure}[b]{0.75\textwidth}
   \includegraphics[width=1\linewidth]{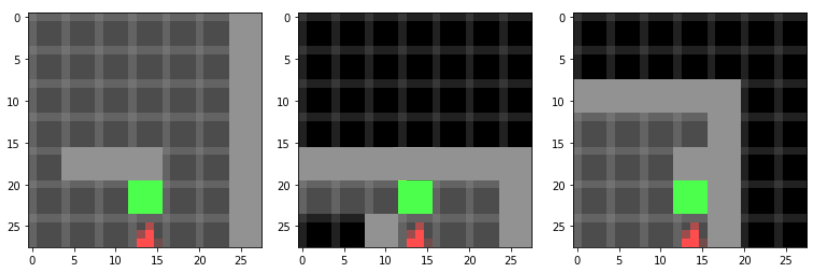}
   \caption{Goal observations.}
   \label{fig:Nw2targets}
\end{subfigure}
  \caption[Example: The two-room environment]{\textbf{Two-room environment.} The red agent must reach the green goal in as few steps as possible. The agent starts each episode between the two rooms, facing a random direction (up, down, left or right).  Each column corresponds to a snapshot of one episode. The light tiles correspond to what the agent sees while the dark tiles are unseen by the agent. (a) Examples of the full state (b) The observation from the agent's current state (c) A goal observation. This is the agent's point of view from a state that separates the agent from the goal by one action. }
  \label{tworooms}
  \end{figure}

The method is tested on three different gridworld environments based on the Minimalistic Gridworld Environment (MiniGrid) \citep{gym_minigrid}. Tiles can be empty, occupied by a wall or occupied by lava. The environments fit naturally into our POMDP tuple template (Eq. \ref{eq:pomdp}): 
\begin{itemize}
    \item[$\mathcal{S}$:]The constituent states of $\mathcal{S}$ are determined by the agent's location and direction (facing north, west, south or east), along with the goal's location. The steps taken since the episode's initialization is also tracked for reward calculations. See Fig. \ref{fig:Nw1b} for three different world states in one of our environments. 
    \item[$\mathcal{A}$:]The action space $\mathcal{A}$ consists of three actions: (1) turn left, (2) turn right and (3) move forward. 
    \item[$\mathcal{P}$:] The transition function is deterministic. The agent relocates to the tile it faces if it moves forward and the tile is empty and nothing happens if the tile is occupied by a wall. The episode terminates if the tile is occupied by lava or the goal. The agent rotates in place if it turns left or right. 
    \item[$\mathcal{R}$:] Reaching the green tile goal gives a reward of $1 - 0.9 \cdot \frac{\# \textrm{steps taken} }{\# \textrm {max step}}/ $, every other action gives 0 points. The environment automatically times out after $\# \textrm {max step}=100$ steps.
    \item[$\mathbb{P}(s_0)$:] Differs between the three environments (see below).
    \item[$\Omega$:] All $7\times7$ subset of tiles, represented by $28\times28\times3$ arrays, from the point of view of an agent who cannot see through walls.
    \item[$O$:] The point of view of the agent from its current viewpoint (Fig. \ref{fig:Nw2obses}) and the goal observation (Fig. \ref{fig:Nw2targets}).
    \item[$\gamma$:] The discount factor is $0.99$.
\end{itemize}

  \subsubsection{Two-room experiment}
  The world is a $8 \times 17$ grid of tiles, split into two rooms (Fig. \ref{tworooms}). The agent is placed between the two rooms, facing a random direction. The goal is at one of three possible locations. This is a modified version of the classical four-room environment layout \citep{sutton1999between}, where walls are placed at different locations to facilitiate discrimination between the rooms from the agent's point of view.

  \subsubsection{Lava gap experiment}

\begin{figure*}[ht]
	\centering
	\begin{minipage}{.79\columnwidth}
		\centering
		\includegraphics[width=\textwidth]{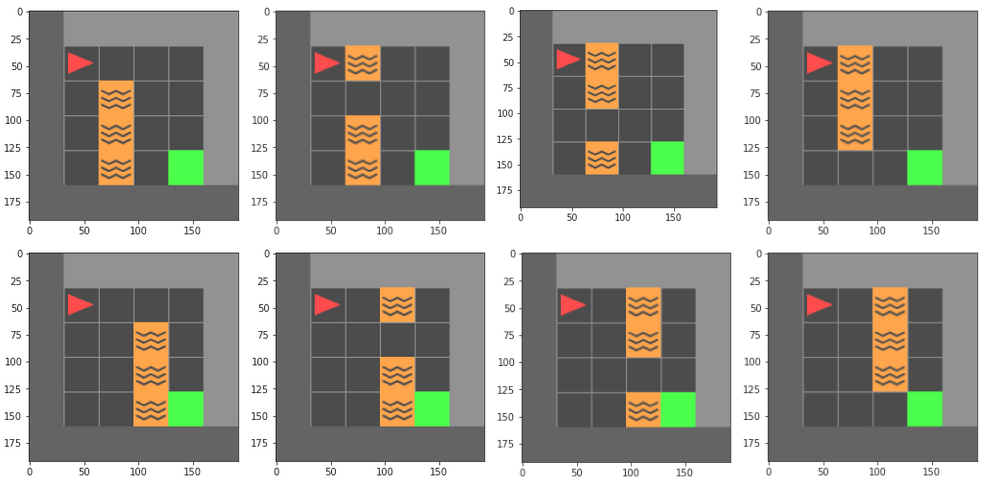}
	\end{minipage}
		\caption[Example: The lava gap environment]{ \textbf{The lava gap environment.} The red agent must reach the green goal in as few steps as possible while avoiding the orange lava tiles. Each episode is randomly selected to be one of the eight pictured configurations. If the agent tries to go through the orange lava tile, it experiences an immediate episode termination with no reward. Note that the wall tiles that are lighter than the others are presently in the agent's field of vision. }
		\label{fig:Ng-1}
\end{figure*}

  In this environment the agent is in a $4 \times 4$ room with a column of lava either one or two spaces in front of the agent (Fig. \ref{fig:Ng-1}) with a gap in a random row. The agent always starts in the upper left corner and the goal is always in the lower right corner.

  \subsubsection{Four-room environment}
\begin{figure*}[ht]
	\centering
	\begin{minipage}{.59\columnwidth}
		\centering
		\includegraphics[width=\textwidth]{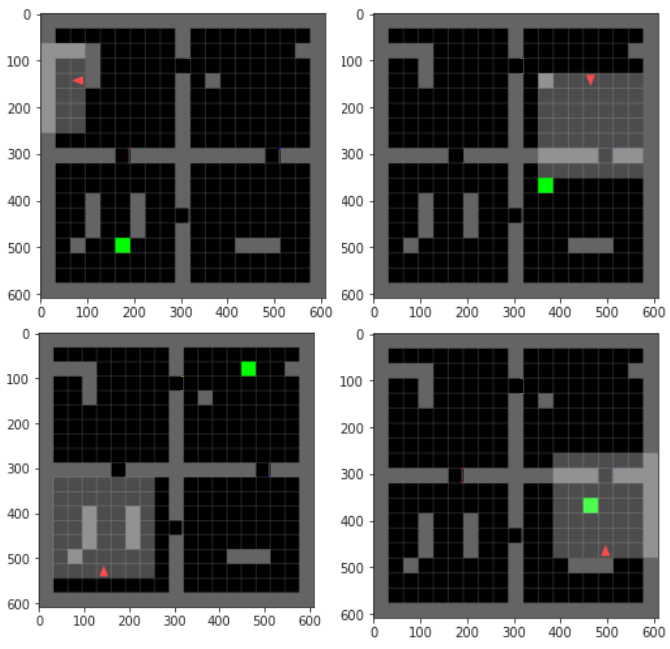}
	\end{minipage}
		\caption[Example: Four-room environment]{ \textbf{Four-room environment}. The agent (red triangle) must reach the goal (green square) in as few steps as possible, both are randomly placed in each episode. The $7\times7$ grid of highlighted tiles in front of the agent indicates its observation.}
		\label{fullillus}
\end{figure*}

  An expansion to the two-room environment with two additional rooms (Fig. \ref{fullillus}). In this setup, both the agent and the goal location are placed at random locations within the $17\times17$ gridworld.
  
\subsection{Baselines} \label{subsec:cm}
 We combine our representations with two RL algorithms as implemented in Stable Baselines \citep{stable-baselines} using the default hyperparameters:

\begin{itemize}

\item (ACKTR) Actor Critic using Kronecker-Factored Trust Region \citep{wu2017scalable}, which combines actor-critic methods, trust-region optimization and distributed Kronecker factorization to enhance sample-efficiency. 
\item (PPO2) A version of the Proximal Policy Optimization algorithm \citep{schulman2017proximal}. It modifies the original algorithm by using clipped value functions and a normalized advantage.
\end{itemize}

For both algorithms, six variations are compared:

\begin{itemize}
    \item (Deep RL) The RL algorithm learns the representation from scratch
    \item (SF) The input is preprocessed using successor features
    \item (Ours 1r) The input is preprocessed using our representation, trained on raw reward predictions
    \item (Ours+Shaping 1r) The input is preprocessed using our representation and the reward is shaped, trained on raw reward predictions
   \item (Ours 64r) The input is preprocessed using our representation, trained on smoothed reward predictions
    \item (Ours+Shaping 1r) The input is preprocessed using our representation and the reward is shaped, trained on smoothed reward predictions

\end{itemize}

Care has been taken to ensure that each variation has the same architecture and the same number of parameters. 

\subsection{Model architectures} \label{subs:ma}

Every model is realized as a neural network using Keras \citep{chollet2015keras}. Below, the representation and policy networks are used for our method and the SF comparison, the reward prediction network is used only for our method and the deep RL network is used only for the deep RL comparison, where the RL algorithm also learns the representation. 

\textbf{The representation networks} are two convolutional networks (Table \ref{tab:repnet})  with a $28 \times 28 \times 3$ input, taking either the agent's current observation or the goal observation.

\begin{table}[ht]
    \centering \begin{tabular}{|l @{\hskip 0.2in} r @{\hskip 0.2in} r @{\hskip 0.2in} l @{\hskip 0.2in} r @{\hskip 0.2in} c|} 
     \hline\rule{0pt}{2.2ex}
     \textbf{Layer} & \textbf{Filters} & \textbf{Filter size} & \textbf{Stride}  & \textbf{Padding} & \textbf{Output shape} \\ [0.5ex] 
     \hline\rule{0pt}{2.2ex}Input tensor & - & -  &-&-&$28\!\times\!28\!\times\!3$  \\[1ex] 
       \hline \rule{0pt}{2.2ex}Convolution & 8 & $3\!\times\!3$  & 3 & None & $9\!\times\!9\!\times\!8$ \\[.5ex] 
      ReLU & - & -  & -&- & $4\!\times\!4\!\times\!8$  \\[.5ex] 
       \hline\rule{0pt}{2.2ex}2D max pooling & 8 & $2\!\times\!2$  & -& None & $4\!\times\!4\!\times\!8$  \\[.5ex] 
     \hline \rule{0pt}{2.2ex}Convolution     & 16 & $3\!\times\!3$  & 2&None & $1\!\times\!1\!\times\!16$   \\[.5ex] 
       ReLU & - & -  & -&-& $1\!\times\!1\!\times\!16$ \\[.5ex] 
       \hline\rule{0pt}{2.2ex}Flatten & - & -  & - &-& 16 \\[.5ex] 
       Dense & - & -  & - &-&16 \\[.5ex]

    \hline
    \end{tabular}
    \vspace{0.1cm}
    \caption[Result: Representation network]{\textbf{Representation network.}}
    \label{tab:repnet}
\end{table}

The first layer subsamples the input, keeping only every other column and row. This is followed by 8 filters of size $3 \times 3$ with a stride of 3. This is followed with a ReLU activation and a $2 \times 2 $ max pooling layer with a stride value of 2. The pooling layer's output is passed to a layer with 16 convolutional filters of size $3 \times 3$ and a stride of 2 and a ReLU activation function. The output is then flattened and passed to a dense layer with 16 units and a linear activation, defining the dimension of the representation. No zero padding is applied in the convolutional layers or the pooling layer.

\textbf{The policy networks} are three-layer fully-connected networks (Table \ref{tab:polnet1}) accepting the concatenated representation of the agent's current point of view and the goal observation as input. The first two layers have 64 units and a ReLU activation and the last layer has 3 units and a linear activation function. 

\begin{table}[ht]
    \centering \begin{tabular}{|l @{\hskip 0.3in} r @{\hskip 0.3in} c|}
     \hline\rule{0pt}{2.2ex}
     \textbf{Layer} & \textbf{Units} &  \textbf{Output shape}   \\ [0.5ex] 
     \hline\rule{0pt}{2.2ex}Input tensor &-&32  \\[1ex] 
     \hline \rule{0pt}{2.2ex}Dense & 64 & 64 \\[.5ex] 
      \rule{0pt}{2.2ex}ReLU & - & 64 \\[.5ex] 

\hline Dense & 64 &3 \\[.5ex] 
ReLU & -  & 3 \\[.5ex] 

\hline Dense & 3 & 3 \\[.5ex] 
    \hline
    \end{tabular}
    \vspace{0.1cm}
    \caption[Result: Policy network]{\textbf{Policy network.}}
    \label{tab:polnet1}
\end{table}

\textbf{Our reward prediction network} is a three-layer fully-connected network (Table \ref{tab:polnet3}) with the same input as the policy network: the concatenated representation of the agent's current view and the goal observation. The first two layers have 256 units and a ReLU activation but the last layer has 1 unit and a logistic activation function. 

\begin{table}[ht]
    \centering \begin{tabular}{|l @{\hskip 0.3in} r @{\hskip 0.3in} c|} 
     \hline\rule{0pt}{2.2ex}
     \textbf{Layer} & \textbf{Units} &  \textbf{Output shape}  \\ [0.5ex] 
     \hline\rule{0pt}{2.2ex}Input tensor &-& 32  \\[1ex] 
     \hline \rule{0pt}{2.2ex}Dense & 256 &64 \\[.5ex] 
      \rule{0pt}{2.2ex}ReLU & - &64 \\[.5ex] 
\hline Dense & 256 & 3 \\[.5ex] 
ReLU & - & 3 \\[.5ex] 
\hline Dense & 1 & 3 \\[.5ex] 
Logistic & - & 3 \\[.5ex] 
    \hline
    \end{tabular}
    \vspace{0.1cm}
    \caption[Result: Reward prediction network]{\textbf{Reward prediction network.}}
    \label{tab:polnet3}
\end{table}

\textbf{The deep RL network} stacks the representation network and the policy network on top of each other. The representation network accepts the input and outputs the low-dimensional representation to the policy network that outputs the action scores.

\subsection{Training the representation and predictor networks} \label{subsec:policy} We collect a data set of $10$ thousand transitions by following a random policy in the two-room environment. For this data collection, each episode has a $50\%$ chance to have the goal location in the bottom room or on the left side of the top room (see the left and middle pictures in Fig. \ref{fig:Nw1b}). The reward predictor and representation are trained in this manner for all experiments, including the lava gap and the four-room environment. The sparse reward associated with the observations in the data set is augmented by associating a new reward to either $1$ (raw reward with no changes made) or $64$ states leading to observations with a positive reward, according to Equation \ref{eq:rstar}, with a discount factor of $0.99$. Additionally, after the reward has been (potenially) smoothed in this way, observations associated with a positive reward are oversampled $10$ times to balance the data set.







\section{Results and discussion}

In the experiments, we compare RL agents that learn their representations from scratch (Deep RL) to preprocessing their inputs with different representations. We compare our representation, trained on raw reward predictions -- with (Ours 1r) or without reward shaping (Ours+Shaping 1r) -- to our method trained on smoothed reward prediction, also with (Ours 64r) or without reward shaping (Ours+Shaping 64r). As a baseline, we also compare our representation to a reward-predictive representation from the literature, Successor Features (SFs). 
\subsection{Two-room environment}



\begin{figure}
\centering
\begin{subfigure}[b]{0.75 \textwidth}
   \includegraphics[width=1\linewidth]{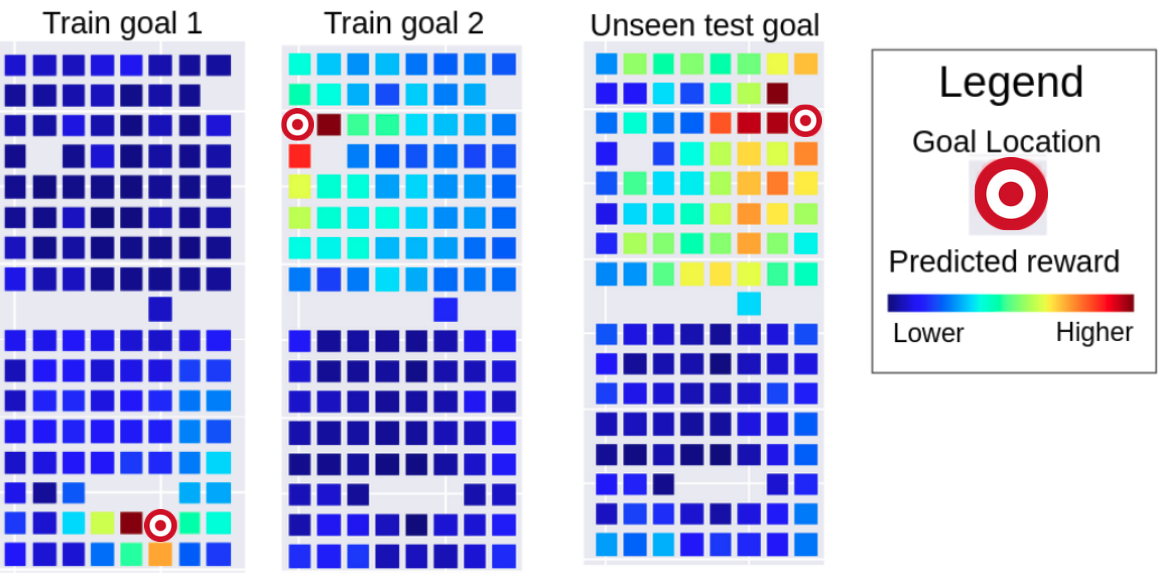}
   \caption{Raw reward prediction.}
   \label{fig:1r}
\end{subfigure}
\begin{subfigure}[b]{0.75\textwidth}
   \includegraphics[width=1\linewidth]{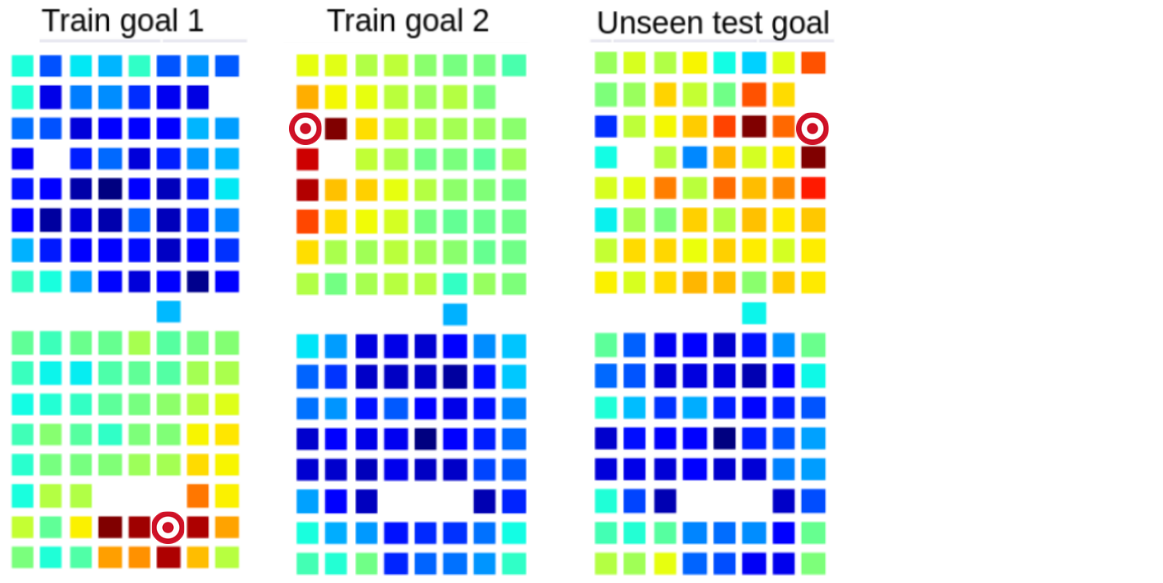}
   \caption{Smoothed reward prediction.}
   \label{fig:64r}
\end{subfigure}
  \caption[Results: Predicted rewards, two-room environment]{\textbf{Predicted rewards, two-room environment.} The predictor is trained on the setups shown on the left and in the middle, and tested for the setup on the right. The color becomes warmer for states where the reward is predicted to be higher.}
  \label{fig:rewardheatmap}
  \end{figure}

We start by visualizing the outputs of our reward predictor in the rooms, depending on the goal location, in Fig. \ref{fig:rewardheatmap}. Each square indicates the average predicted reward for transitioning to the corresponding tile in the room. 

The reward spikes in a narrow region around the two goal locations
that were used to train the raw reward predictor (Fig. \ref{fig:1r}), but the area of states with high predicted rewards is
wider around the test goal. This difference is due to overfitting on the specific training paths that
were more frequently taken toward the respective goals, but this does not harm the generalization
capabilities of the network. The peakyness of the predictions disappears when the predictor is trained on the smoothed rewards (Fig. \ref{fig:64r}). However, much higher predicted rewards in the corner of the other room appears. Both scenarios, raw and smoothed reward prediction, show promise for the application of reward shaping under our training scheme, as the agent would benefit from find neighborhoods with higher values of predicted reward until it reaches the goal, instead of having to rely solely on a sparse reward that is only given when the agent lands exactly on the goal state.

\begin{figure*}[ht]
	\centering
	\begin{minipage}{.99\columnwidth}
		\centering
		\includegraphics[width=\textwidth]{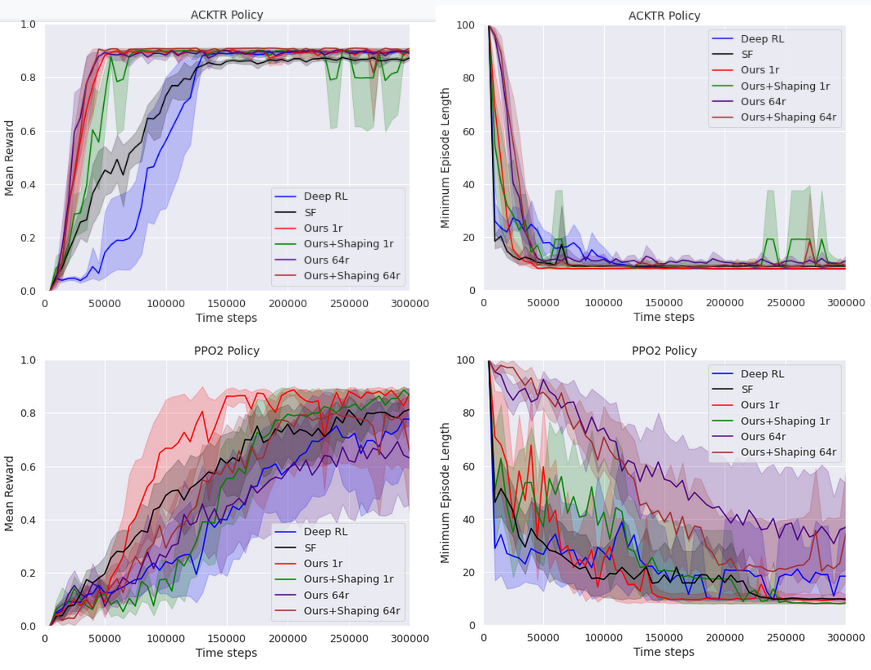}
	\end{minipage}
		\caption[Result: Single goal location]{ \textbf{Single goal location}. In these experiments, there are only two rooms the agent must reach a goal that is always at the same location. The agent can traverse between the rooms and starts each episode between them, facing a random direction. }
		\label{allsingle}
\end{figure*}

In Fig. \ref{allsingle}, we illustrate the mean reward of the different methods, as a function of the time steps taken for training, as well as the minimum episode lengths. We average over 10 runs and in each run we perform 10 test rollouts, so each point is the aggregate of 100 episodes in total. The error bands indicate two standard deviations. Note that this methodology of generating the plots also applies to Fig. \ref{fig:fromscratch} and Fig. \ref{allfull2}.

The learning curves of both ACKTR and PPO2 get close to the highest achievable mean reward of $1$ the fastest using our representations. There no significant benefit from using smoothed reward shaping for ACKTR, and the raw reward shaping is in fact harmful in this case. For PPO2, the agent using our representation that is trained on raw reward predictions learns the fastest. Regular deep RL, where the representations are learned from scratch, is clearly outperformed by the variants that use reward-predictive representations. We believe that this is because RL agents can generally benefit from the input being preprocessed, as the computational overhead for learning the policy is reduced. This effect is enhanced when the preprocessing is good, which is the case for our reward-predictive representation: it abstracts away unnecessary information as it trained to output features that indicate the distance between the agent and the goal, when the goal is in view.

\newpage

\subsection{Lava gap environment}

  \subsubsection{Learning from scratch}

\begin{figure*}[ht]
	\centering
	\begin{minipage}{.37\columnwidth}
		\centering
		\includegraphics[width=\textwidth]{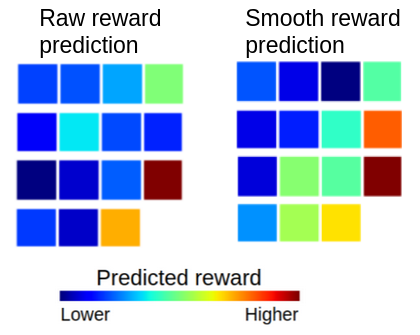}
	\end{minipage}
		\caption[Result: Predicted rewards, lava gap]{ \textbf{Predicted rewards, lava gap}. Average predicted reward per state in the lava gap environment. }
		\label{fig:gapheatmap}
\end{figure*}

\begin{figure*}[ht]
	\centering
	\begin{minipage}{.99\columnwidth}
		\centering
		\includegraphics[width=\textwidth]{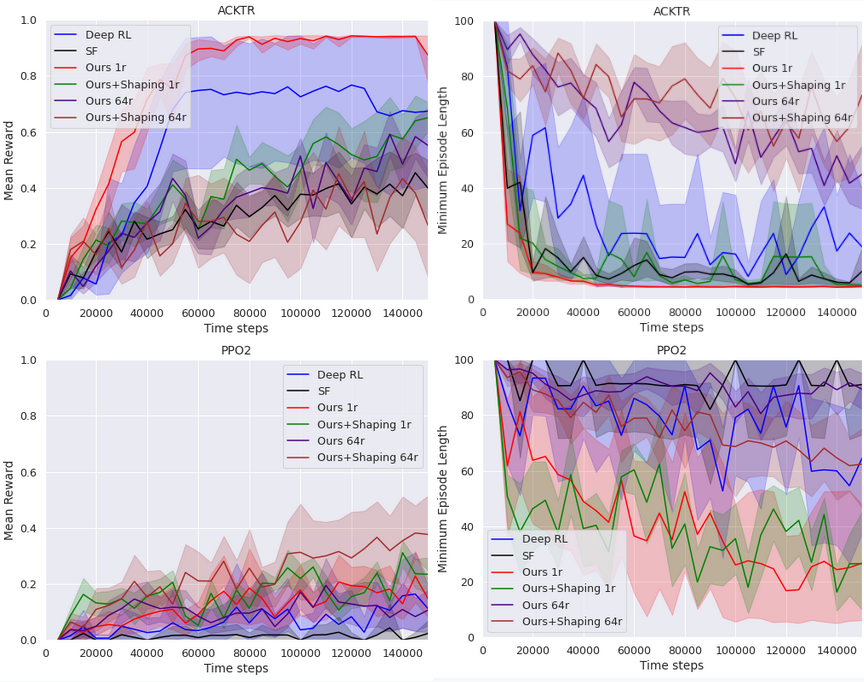}
	\end{minipage}
		\caption[Result: Lava gap experiment]{ \textbf{Lava gap experiment.} All policies are randomly initialized and learn to solve the lava gap environment from scratch. The representations in all methods except for Deep RL are learned on the training goals in the two-room experiment (see Fig. \ref{fig:rewardheatmap})  }
		\label{fig:fromscratch}
\end{figure*}

The heatmap of average predicted rewards are visualized in Fig. \ref{fig:gapheatmap}. The reward predictor was trained on the two-goal environment. The tiles closest to the goal have the highest values, with a particularly smooth gradient toward the goal for the smoothed-reward predictor, which demonstrates that there is potential gain from transfering the prediction-based reward shaping between similar environments. The learning performance of the different methods can be seen in Fig. \ref{fig:fromscratch}. The decidedly fastest learning can be observed when the actor-critic method is combined with our representation, trained on raw reward predictions and without reward shaping. Regular deep RL is the second-best but with a very large variance on the performance. Our reward shaping variations and the SFs are very close in performance, albeit significantly worse than the other two. The poor performance of reward shaping can be explained by the fact that there are very few states, which makes the reward shaping unnecessary in such a simple environment.  All the methods look more similar when PPO2 optimization is applied, with respect to the mean rewards, but our variant that is trained on smooth reward prediction and uses reward shaping reaches the highest average performance in the last iterations.

 \subsubsection{Transfer learning}
 
\begin{figure*}[ht]
	\centering
	\begin{minipage}{.99\columnwidth}
		\centering
		\includegraphics[width=\textwidth]{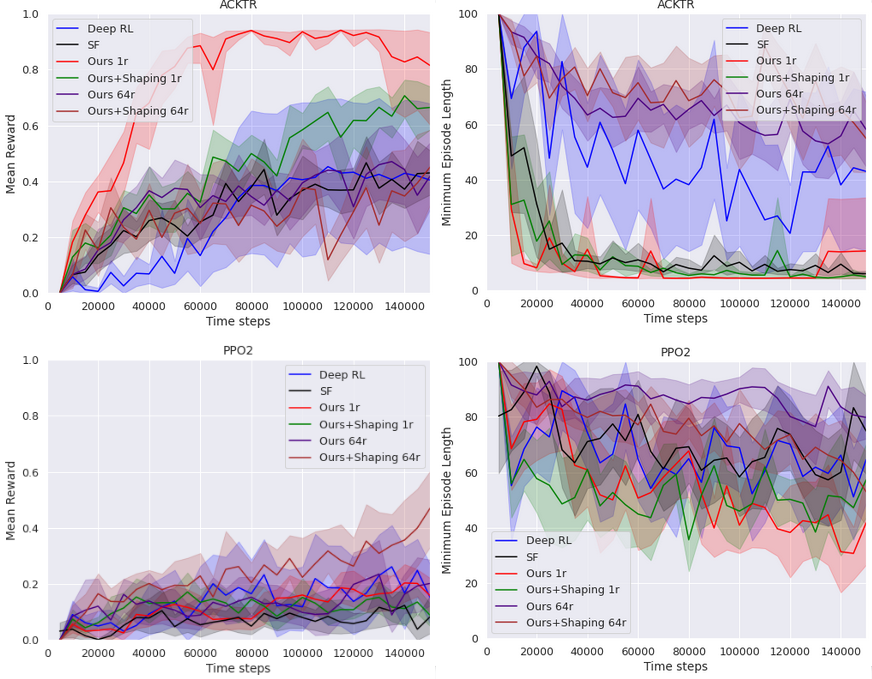}
	\end{minipage}
		\caption[Result: Re-learning experiment]{ \textbf{Re-learning experiment} The different methods are trained for eight thousand training steps on the two-room environment before being trained on the lava gap environment. The curves show the mean reward on the lava gap environment.}
		\label{graph:8k}
\end{figure*}

 To investigate how the methods compare for adapting to new environments, we trained all methods for 8000  steps on the two-room environment before learning to solve the lava gap environment, see Fig. \ref{graph:8k}. Our method, without reward shaping, facilitates the fastest learning for ACKTR in this case. Deep RL is the most severely affected by this change, which is probably due to the method learning a reward-maximizing representation in one environment that does not transfer well to another environment. Every PPO2 variation looks bad for this scenario, but smooth-reward prediction representation with reward shaping has the highest mean reward and our raw-reward prediction representation has the lowest average minimum episode length. 

\begin{figure}[bht]
\centering
\begin{subfigure}[b]{0.47\textwidth}
   \includegraphics[width=1\linewidth]{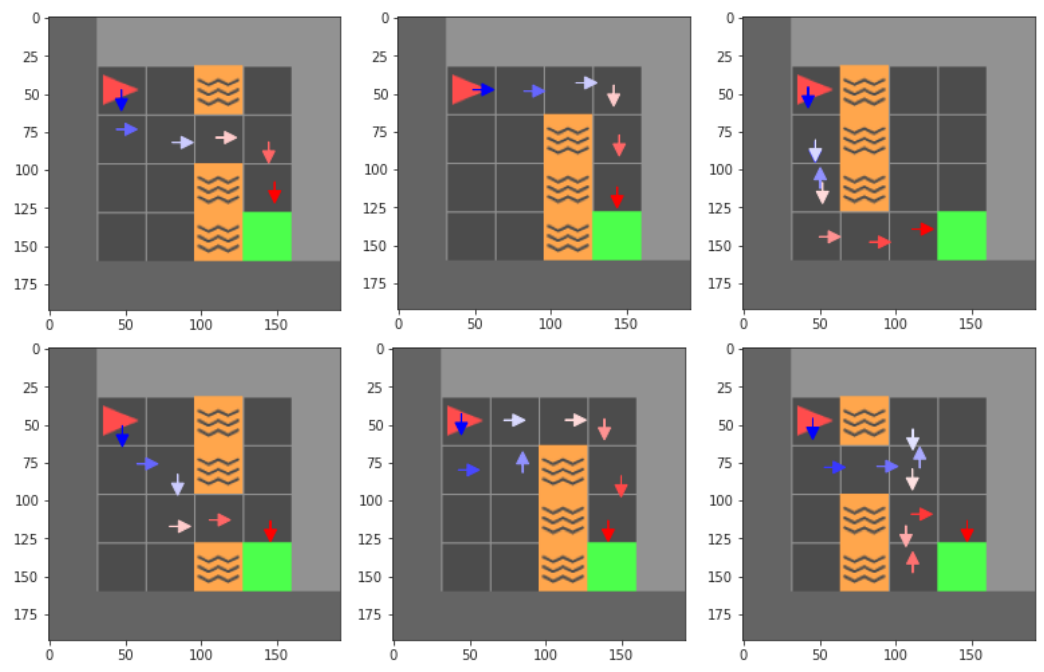}
   \caption{Six successful episodes.}
   \label{fig:Ng1} 
\end{subfigure}
\begin{subfigure}[b]{0.47\textwidth}
   \includegraphics[width=1\linewidth]{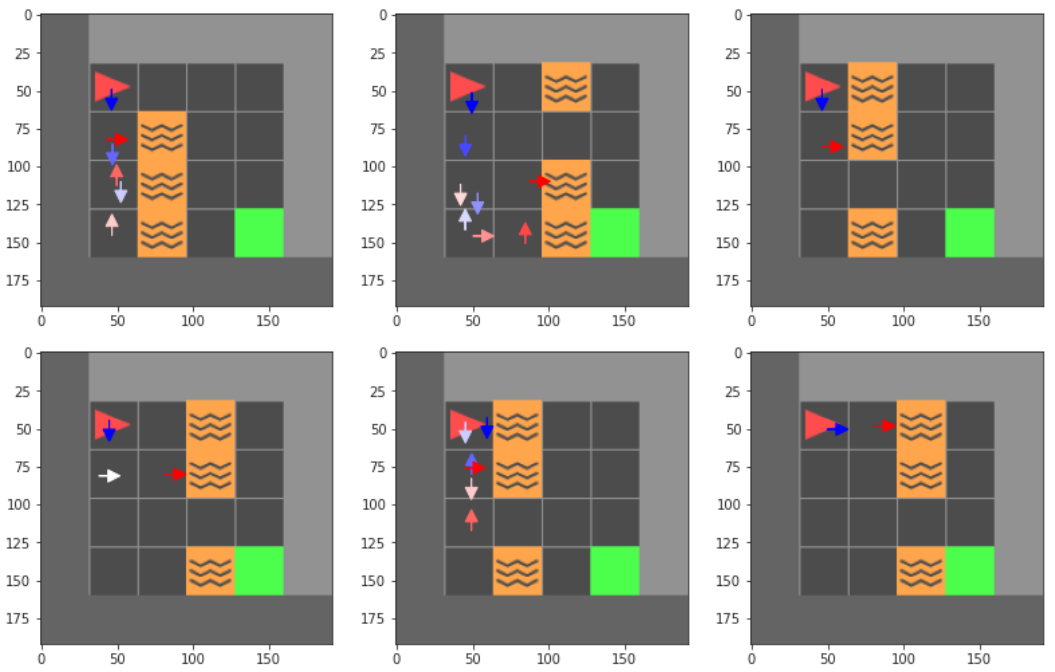}
   \caption{Six failed episodes.}
   \label{fig:Ng2}
\end{subfigure}
\caption[Results: Lava gap trajectories]{\textbf{Lava gap trajectories.} A visualization of six successful and six failed trajectories by an agent that was trained up to approximately $75\%$ success rate using our representation. The color gradient goes from the first actions taken in the episode in blue to the last actions in red. Note that rotations in-place are not visualized but only transitions between tiles. }
\label{lavatraj}
\end{figure}

We visualize trajectories of an agent that's trained on our representation (Ours 1r) as it traverses the lava gaps environment (Fig. \ref{lavatraj}). For inspection of cases where it fails, we choose an agent that has been trained for $50$ thousand time steps only and around $0.75$ mean reward.

\subsection{Four-room environment}

\begin{figure*}[ht]
	\centering
	\begin{minipage}{.97\columnwidth}
		\centering
		\includegraphics[width=\textwidth]{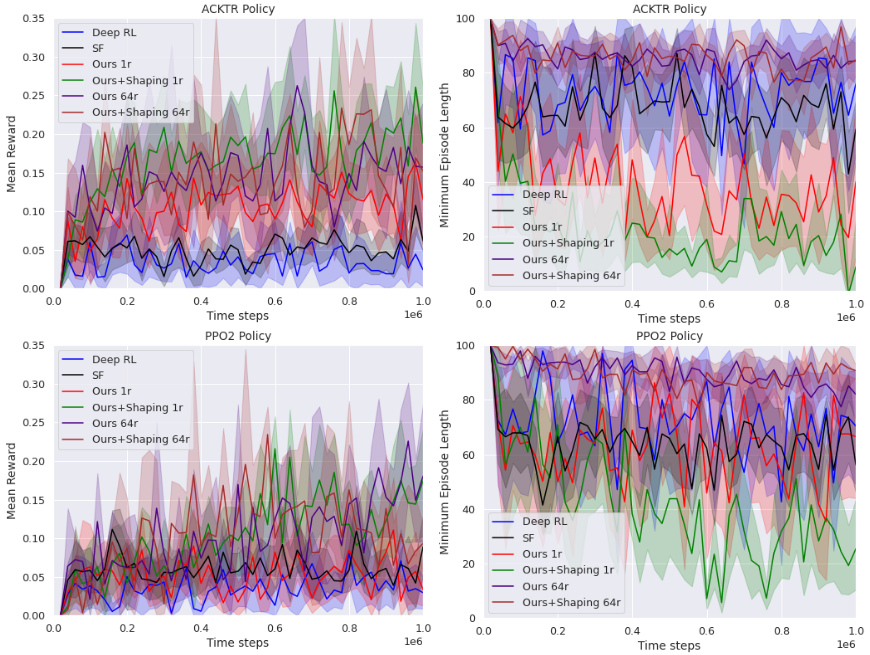}
	\end{minipage}
		\caption[Result: Full four-room experiment]{ \textbf{Full four-room experiment}. The agent and goal locations both start each episode at random locations.}
		\label{allfull2}
\end{figure*}

In our final comparison, we add two additional rooms to the two-room environment and randomize both the goal location and the starting position of the agent, with the results visible in Fig. \ref{allfull2}.  Looking at the minimum episode lengths, for the ACKTR learner, our raw-reward prediction representation with reward shaping performs the best and our representation without reward shaping comes in second.  There is little discernible difference between the performance of SFs and Deep RL, but they both perform significantly worse than our methods. The scale of the mean reward is a great deal lower in the previous experiments since the average distance between the starting tile of the agent and the goal is much larger than in the previous two environments. 

For this scenario, all the methods look similarly bad for the PPO2 policy, except for our raw-reward representations with reward shaping which has the lowest minimum episode length. The big difference between the performance of reward shaping in this environment and the two-room environment can be explained by the increased complexity, making the reward shaping more helpful in guiding the agent's search. In the previous experiments, the agent and goal locations start at fixed locations, allowing the agents to solve it by rote memorization.  The reward shaping function calculated by the raw-reward predictor fares significantly better in this situation. We hypothesize that this is due to the smoothed-reward predictor distracting the agent by pushing it to corners, as the visualization in Fig. \ref{fig:64r} would suggest. 

The reward shaping given by the raw-reward predictor is more discriminative, as we see in Fig. \ref{fig:1r}. The agent receives a positive reward as soon as a goal reaches its point of view, which is any location up to six tiles in front of it and no further than 3 tiles away from it to the left or to the right. This allows the reward shaping function to guide the agent directly to the goal, assuming that they are in the same room and that there is no wall obstructing the agent's field of vision.

\begin{figure}
\centering
\begin{subfigure}[b]{0.75\textwidth}
   \includegraphics[width=1\linewidth]{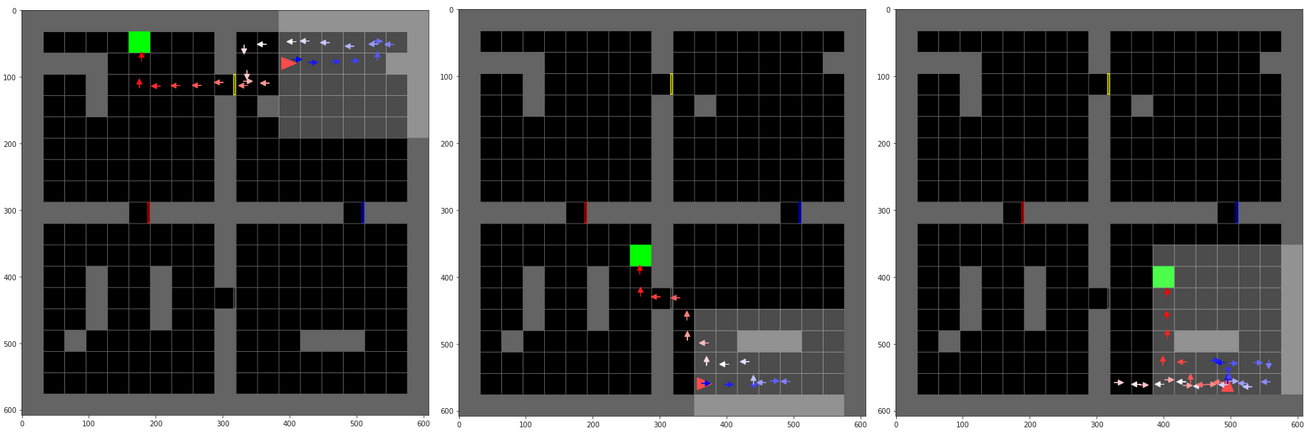}
   \caption{Three successful trajectories}
   \label{fig:Ng3} 
\end{subfigure}
\begin{subfigure}[b]{0.75\textwidth}
   \includegraphics[width=1\linewidth]{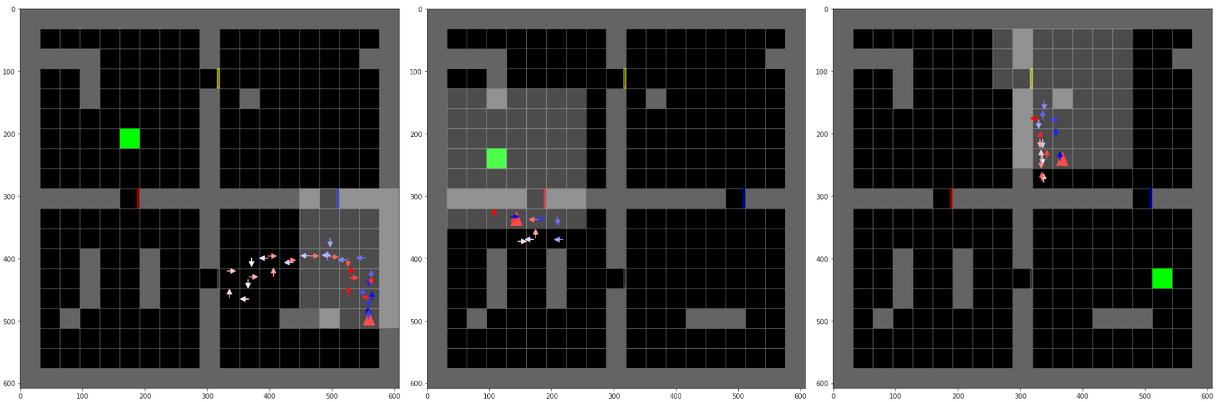}
   \caption{Three failed trajectories}
   \label{fig:Ng4}
\end{subfigure}
\caption[Results: Four-room trajectories]{\textbf{Four-room trajectories.} The agent was trained on our representations with Actor Critic using Kronecker-Factored Trust Region for a million time steps. }
\label{viz4rooms}
\end{figure}

Three successful and three failed trajectories of an AKCTR agent that has been trained, using our representation (Ours 1r), for a million time steps, is visualized in Fig. \ref{viz4rooms}.  We can see the undesirable behavior in the both the succesful and the failed trajectories that the agent wastes effort re-visiting tiles it has already been to.

\section{Conclusion} \label{sec:disc}

 Processing high-dimensional inputs for reinforcement learning agents remains a difficult problem, especially if the agent must rely on a sparse reward signal to guide its representation learning. In this work, we put forward a method to help alleviate this problem with a method of learning representations that preprocess visual input for reinforcement learning (RL) methods. Our contributions are (i) a reward-predictive representation that is trained  simultaneously with a reward predictor and (ii) a reward shaping technique using this trained predictor. The predictor learns to approximate either the raw reward signal or a smoothed version of it and it is used for reward shaping by encouraging the agent to transition to states with higher predicted rewards.
 
 We have shown the usefulness of our representation and our reward shaping scheme in a series of gridworld experiments, where the agent receives a high-dimensional observation of its goal as input along with an observation of its immediate surroundings. Preprocessing the input using this representation speeds up the training of two out-of-the-box RL methods: Actor Critic using Kronecker-Factored Trust Region and Proximal Policy Optimization, compared to having these methods learn the representations from scratch. In our most complicated experiment, combining our representation with our reward shaping technique is shown to perform significantly better than the vanilla RL methods, which hints at its potential for success, especially in more complex RL scenarios.
 
 \section{Acknowledgement}
 
 We would like to thank Moritz Lange for his valuable contribution to the preparation of this paper.
 
 \clearpage

\pdfbookmark[1]{Bibliography}{bibliography}
\bibliography{main}

\begin{thebibliography}{27}
\providecommand{\natexlab}[1]{#1}
\providecommand{\url}[1]{\texttt{#1}}
\expandafter\ifx\csname urlstyle\endcsname\relax
  \providecommand{\doi}[1]{doi: #1}\else
  \providecommand{\doi}{doi: \begingroup \urlstyle{rm}\Url}\fi

\bibitem[Amodei and Clark(2016)]{clark2016faulty}
Dario Amodei and Jack Clark.
\newblock Faulty reward functions in the wild, 2016.
\newblock \url{https://blog.openai.com/faulty-reward-functions/, 2016}.

\bibitem[Barreto et~al.(2016)Barreto, Dabney, Munos, Hunt, Schaul, Van~Hasselt,
  and Silver]{barreto2016successor}
Andr{\'e} Barreto, Will Dabney, R{\'e}mi Munos, Jonathan~J Hunt, Tom Schaul,
  Hado Van~Hasselt, and David Silver.
\newblock Successor features for transfer in reinforcement learning.
\newblock \emph{arXiv preprint arXiv:1606.05312}, 2016.

\bibitem[Berner et~al.(2019)Berner, Brockman, Chan, Cheung, D{\k{e}}biak,
  Dennison, Farhi, Fischer, Hashme, Hesse, et~al.]{berner2019dota}
Christopher Berner, Greg Brockman, Brooke Chan, Vicki Cheung, Przemys{\l}aw
  D{\k{e}}biak, Christy Dennison, David Farhi, Quirin Fischer, Shariq Hashme,
  Chris Hesse, et~al.
\newblock Dota 2 with large scale deep reinforcement learning.
\newblock \emph{arXiv preprint arXiv:1912.06680}, 2019.

\bibitem[Brown and Sandholm(2019)]{brown2019superhuman}
Noam Brown and Tuomas Sandholm.
\newblock Superhuman ai for multiplayer poker.
\newblock \emph{Science}, 365\penalty0 (6456):\penalty0 885--890, 2019.

\bibitem[Brys et~al.(2014)Brys, Harutyunyan, Vrancx, Taylor, Kudenko, and
  Now{\'e}]{brys2014multi}
Tim Brys, Anna Harutyunyan, Peter Vrancx, Matthew~E Taylor, Daniel Kudenko, and
  Ann Now{\'e}.
\newblock Multi-objectivization of reinforcement learning problems by reward
  shaping.
\newblock In \emph{2014 international joint conference on neural networks
  (IJCNN)}, pages 2315--2322. IEEE, 2014.

\bibitem[Brys et~al.(2015)Brys, Harutyunyan, Taylor, and
  Now{\'e}]{brys2015policy}
Tim Brys, Anna Harutyunyan, Matthew~E Taylor, and Ann Now{\'e}.
\newblock Policy transfer using reward shaping.
\newblock In \emph{Proceedings of the 2015 International Conference on
  Autonomous Agents and Multiagent Systems}, pages 181--188, 2015.

\bibitem[Chevalier-Boisvert et~al.(2018)Chevalier-Boisvert, Willems, and
  Pal]{gym_minigrid}
Maxime Chevalier-Boisvert, Lucas Willems, and Suman Pal.
\newblock Minimalistic gridworld environment for openai gym.
\newblock \url{https://github.com/maximecb/gym-minigrid}, 2018.

\bibitem[Chollet et~al.(2015)]{chollet2015keras}
Fran\c{c}ois Chollet et~al.
\newblock Keras.
\newblock https://keras.io, 2015.

\bibitem[Efthymiadis and Kudenko(2013)]{efthymiadis2013using}
Kyriakos Efthymiadis and Daniel Kudenko.
\newblock Using plan-based reward shaping to learn strategies in starcraft:
  Broodwar.
\newblock In \emph{2013 IEEE Conference on Computational Inteligence in Games
  (CIG)}, pages 1--8. IEEE, 2013.

\bibitem[Hill et~al.(2018)Hill, Raffin, Ernestus, Gleave, Kanervisto, Traore,
  Dhariwal, Hesse, Klimov, Nichol, Plappert, Radford, Schulman, Sidor, and
  Wu]{stable-baselines}
Ashley Hill, Antonin Raffin, Maximilian Ernestus, Adam Gleave, Anssi
  Kanervisto, Rene Traore, Prafulla Dhariwal, Christopher Hesse, Oleg Klimov,
  Alex Nichol, Matthias Plappert, Alec Radford, John Schulman, Szymon Sidor,
  and Yuhuai Wu.
\newblock Stable baselines.
\newblock \url{https://github.com/hill-a/stable-baselines}, 2018.

\bibitem[Kaelbling(1993)]{kaelbling1993learning}
Leslie~Pack Kaelbling.
\newblock Learning to achieve goals.
\newblock In \emph{IJCAI}, pages 1094--1099. Citeseer, 1993.

\bibitem[Lample and Chaplot(2017)]{lample2017playing}
Guillaume Lample and Devendra~Singh Chaplot.
\newblock Playing fps games with deep reinforcement learning.
\newblock \emph{Proceedings of the AAAI Conference on Artificial Intelligence},
  31\penalty0 (1), 2017.

\bibitem[Lehnert and Littman(2020)]{lehnert2020successor}
Lucas Lehnert and Michael~L Littman.
\newblock Successor features combine elements of model-free and model-based
  reinforcement learning.
\newblock \emph{Journal of Machine Learning Research}, 21\penalty0
  (196):\penalty0 1--53, 2020.

\bibitem[Lehnert et~al.(2020)Lehnert, Littman, and Frank]{lehnert2020reward}
Lucas Lehnert, Michael~L Littman, and Michael~J Frank.
\newblock Reward-predictive representations generalize across tasks in
  reinforcement learning.
\newblock \emph{PLoS computational biology}, 16\penalty0 (10):\penalty0
  e1008317, 2020.

\bibitem[Marashi et~al.(2012)Marashi, Khalilian, and
  Shiri]{marashi2012automatic}
Maryam Marashi, Alireza Khalilian, and Mohammad~Ebrahim Shiri.
\newblock Automatic reward shaping in reinforcement learning using graph
  analysis.
\newblock In \emph{2012 2nd International eConference on Computer and Knowledge
  Engineering (ICCKE)}, pages 111--116. IEEE, 2012.

\bibitem[Mataric(1994)]{mataric1994reward}
Maja~J Mataric.
\newblock Reward functions for accelerated learning.
\newblock In \emph{Machine learning proceedings 1994}, pages 181--189.
  Elsevier, 1994.

\bibitem[Mnih et~al.(2013)Mnih, Kavukcuoglu, Silver, Graves, Antonoglou,
  Wierstra, and Riedmiller]{mnih2013playing}
Volodymyr Mnih, Koray Kavukcuoglu, David Silver, Alex Graves, Ioannis
  Antonoglou, Daan Wierstra, and Martin Riedmiller.
\newblock Playing atari with deep reinforcement learning.
\newblock \emph{arXiv preprint arXiv:1312.5602}, 2013.

\bibitem[Ng et~al.(1999)Ng, Harada, and Russell]{ng1999policy}
Andrew~Y Ng, Daishi Harada, and Stuart Russell.
\newblock Policy invariance under reward transformations: Theory and
  application to reward shaping.
\newblock In \emph{Icml}, volume~99, pages 278--287, 1999.

\bibitem[Pathak et~al.(2018)Pathak, Mahmoudieh, Luo, Agrawal, Chen, Shentu,
  Shelhamer, Malik, Efros, and Darrell]{pathak2018zero}
Deepak Pathak, Parsa Mahmoudieh, Guanghao Luo, Pulkit Agrawal, Dian Chen, Yide
  Shentu, Evan Shelhamer, Jitendra Malik, Alexei~A Efros, and Trevor Darrell.
\newblock Zero-shot visual imitation.
\newblock In \emph{Proceedings of the IEEE Conference on Computer Vision and
  Pattern Recognition Workshops}, pages 2050--2053, 2018.

\bibitem[Schaul et~al.(2015)Schaul, Horgan, Gregor, and
  Silver]{schaul2015universal}
Tom Schaul, Daniel Horgan, Karol Gregor, and David Silver.
\newblock Universal value function approximators.
\newblock In \emph{International conference on machine learning}, pages
  1312--1320, 2015.

\bibitem[Schulman et~al.(2015)Schulman, Moritz, Levine, Jordan, and
  Abbeel]{schulman2015high}
John Schulman, Philipp Moritz, Sergey Levine, Michael Jordan, and Pieter
  Abbeel.
\newblock High-dimensional continuous control using generalized advantage
  estimation.
\newblock \emph{arXiv preprint arXiv:1506.02438}, 2015.

\bibitem[Schulman et~al.(2017)Schulman, Wolski, Dhariwal, Radford, and
  Klimov]{schulman2017proximal}
John Schulman, Filip Wolski, Prafulla Dhariwal, Alec Radford, and Oleg Klimov.
\newblock Proximal policy optimization algorithms.
\newblock \emph{arXiv preprint arXiv:1707.06347}, 2017.

\bibitem[Silver et~al.(2016)Silver, Huang, Maddison, Guez, Sifre, Van
  Den~Driessche, Schrittwieser, Antonoglou, Panneershelvam, Lanctot,
  et~al.]{silver2016mastering}
David Silver, Aja Huang, Chris~J Maddison, Arthur Guez, Laurent Sifre, George
  Van Den~Driessche, Julian Schrittwieser, Ioannis Antonoglou, Veda
  Panneershelvam, Marc Lanctot, et~al.
\newblock Mastering the game of go with deep neural networks and tree search.
\newblock \emph{nature}, 529\penalty0 (7587):\penalty0 484--489, 2016.

\bibitem[Sutton et~al.(1999)Sutton, Precup, and Singh]{sutton1999between}
Richard~S Sutton, Doina Precup, and Satinder Singh.
\newblock Between mdps and semi-mdps: A framework for temporal abstraction in
  reinforcement learning.
\newblock \emph{Artificial intelligence}, 112\penalty0 (1-2):\penalty0
  181--211, 1999.

\bibitem[Trott et~al.(2019)Trott, Zheng, Xiong, and Socher]{trott2019keeping}
Alexander Trott, Stephan Zheng, Caiming Xiong, and Richard Socher.
\newblock Keeping your distance: Solving sparse reward tasks using
  self-balancing shaped rewards.
\newblock \emph{arXiv preprint arXiv:1911.01417}, 2019.

\bibitem[Wu et~al.(2017)Wu, Mansimov, Grosse, Liao, and Ba]{wu2017scalable}
Yuhuai Wu, Elman Mansimov, Roger~B Grosse, Shun Liao, and Jimmy Ba.
\newblock Scalable trust-region method for deep reinforcement learning using
  kronecker-factored approximation.
\newblock In \emph{Advances in neural information processing systems}, pages
  5279--5288, 2017.

\bibitem[Zou et~al.(2019)Zou, Ren, Yan, Su, and Zhu]{zou2019reward}
Haosheng Zou, Tongzheng Ren, Dong Yan, Hang Su, and Jun Zhu.
\newblock Reward shaping via meta-learning.
\newblock \emph{arXiv preprint arXiv:1901.09330}, 2019.

\end{thebibliography}
\clearpage

\end{document}